# Accurate Remaining Useful Life Prediction with Uncertainty Quantification: a Deep Learning and Nonstationary Gaussian Process Approach

Zhaoyi Xu, Yanjie Guo, Joseph Homer Saleh

*School of Aerospace Engineering, Georgia Institute of Technology, Atlanta, GA 30332, USA*

**Abstract:** Remaining useful life (RUL) refers to the expected remaining lifespan of a component or system. Accurate RUL prediction is critical for prognostic and health management and for maintenance planning. In this work, we address three prevalent challenges in data-driven RUL prediction, namely the handling of high dimensional input features, the robustness to noise in sensor data and prognostic datasets, and the capturing of the time-dependency between system degradation and RUL prediction. We devise a highly accurate RUL prediction model with uncertainty quantification, which integrates and leverages the advantages of deep learning and nonstationary Gaussian process regression (DL-NSGPR). We examine and benchmark our model against other advanced data-driven RUL prediction models using the turbofan engine dataset from the NASA prognostic repository. Our computational experiments show that the DL-NSGPR predictions are highly accurate with root mean square error 1.7 to 6.2 times smaller than those of competing RUL models. Furthermore, the results demonstrate that RUL uncertainty bounds with the proposed DL-NSGPR are both valid and significantly tighter than other stochastic RUL prediction models. We unpack and discuss the reasons for this excellent performance of the DL-NSGPR.

# I. INTRODUCTION

The concept of remaining useful life (RUL) refers to the expected remaining lifespan of a component or system. Its prediction is used to inform maintenance decisions and to minimize the risk of catastrophic failures during operation [1]. Accurate RUL prediction is critical for prognostic and health management (PHM) and for maintenance planning. Preventive maintenance actions are scheduled based on RUL predictions and equipment's health condition to prevent unexpected failures and reduce total maintenance costs [2]. Accurate RUL prediction is critical for effective condition-based maintenance (CBM), and to sustain equipment reliability and reduce overall maintenance costs [3].

With the significant advancement in sensor technologies and data acquisition in the last decade, data-driven methods for RUL prediction have become increasingly popular [4-12], and different machine learning RUL models have been proposed in a host of applications [13, 14]. For example, Liao and Kottig [15] developed a hybrid Bayesian model for RUL prediction and reported high level of accuracy compared with traditional Bayesian models. Zio and Maio [16] proposed a similar data-driven approach for online RUL estimation. Si et al. [17] provide a comprehensive review of data-driven approaches to RUL prediction.

Despite the recognition of its importance and progress in its prediction, three significant challenges remain in relation to data-driven RUL models. The first challenge is related to the high dimensionality of the datasets, which makes it difficult to extract meaningful features from massive inputs with predictive value for RUL estimation. High dimensional inputs are ever more prevalent, and they increase the complexity of data-driven RUL models. For example, in the turbofan engine prognostic dataset used in this work (details in Section III), the model input is a 24-dimension features vector, and it is not uncommon to find engineering systems with significantly higher dimensional feature vectors. The second challenge is related to the fact that in real-world applications, sensor measurements are often noisy or corrupted, and the uncertainty associated with the RUL prediction increases with sensor noise and can become too large to meaningfully inform maintenance planning. The third challenge is a bit more subtle to appreciate, and it has both theoretical and practical underpinnings. It relates to the time-dependence between the system degradation and RUL prediction given a suite of sensors data. Current prognostic models make RUL predictions based on the present condition and sensor measurements without accounting for the history of operation of the system [1, 2, 18]. Sensor measurements, however, rarely if ever capture the entirety of the state vector of a system, and as a result, two similar measurements from the suite of sensors, for example, arrived at through different trajectories or degradation paths, do not necessarily reflect the same underlying health condition of the system (competing risks and different failure modes can exhibit different degradation paths). Making RUL predictions that do not account for time-dependence in sensor measurements can be variably inaccurate and misinform maintenance decisions, which in turn can fail to prevent run-to-failures (more details in the next section).

We propose in this work to jointly address these three challenges in RUL prediction. However, before we discuss our approach, it is important to acknowledge that this is a vigorous research area and several approaches have been proposed to tackle one or more of these challenges, usually in isolation. We provide next a cursory overview of this literature. There is a broader context within which our work on RUL prediction is situated. It is related to advances in Machine Learning (ML) in general, in Deep Learning (DL) and Gaussian Processes in particular for reliability and safety applications. We recently provided a review of this broad analytical landscape, and we include here a short excerpt for the convenience of the reader. More details can be found in [19]. Among various ML technologies, deep learning (DL) is a key enabler of applications and analysis involving high dimensional data. It uses deep multi-layered neural architecture to fulfill complex regression, classification, and some unsupervised functions. DL can improve model accuracy and tease out more latent information from the raw data compared with shallow ML methods. DL is increasingly used in the analysis of system degradation and RUL prediction in large part because of its superior ability to handle high-dimensional data. For example, Fink et al. [20] devised a multilayer feedforward neural network for RUL and reliability prediction of complex systems and reported high-level accuracy in their validation tests using real-world railway data. Similarly, Li et al. [21] proposed a DL based RUL estimation model and applied it to a bearing prognostic dataset. They also reported high accuracy in their RUL prediction using the bearing prognostic dataset. Another important tool that is used for RUL prediction is Gaussian process regression (GPR). GPR is a nonparametric, statistical, and Bayesian approach to model the evolution of time streaming data with an uncertainty quantification over time. For example, Lederer et al. [22] developed a real-time GPR model for online RUL prediction and reported better accuracy compared with other models. A more extensive review of different ML approaches for RUL prediction can be found in [19]. Another increasingly popular RUL prediction approach focuses on modelling a composite health index of a system by data fusion method [23-27] to address the high-dimensional input challenge. For example, Wang et. al. [23] developed a DL based HI estimation and prognostic model and reported a high-level of prediction accuracy. While highly promising, these approaches do not jointly address the three challenges noted previously, in particular the time-dependence in sensor data, which is often a main cause of large inaccuracies in RUL prediction.

Our objective of this work is to develop a highly accurate RUL prediction model that effectively addresses collectively these three challenges. To this end, we combine and adopt the advantages of the DL neural network for high dimensional dataset analysis with those of GPR, in particular non-stationary GPR [28, 29] to improve the model accuracy and provide uncertainty quantification around the RUL prediction. Our proposed approach is an integrated DL and nonstationary GPR stochastic RUL prediction model (mean and standard deviation).

The main contribution of this work is the development and validation of a highly accurate DL-NSGPR prediction model of RUL with uncertainty quantification. Our method adopts the advantages of DL neural network and non-stationary GPR to jointly address three challenges in the RUL prediction discussed previously. Our RUL prediction is tested and benchmarked using the engine prognostic dataset (C-MAPSS), and results show higher accuracy and tighter uncertainty bounds of our model compared with other best-in-class alternatives in the literature.

The remainder of the article is organized as follows. Our RUL prediction model and its technical details are introduced in Section II. The datasets and computational experiments used to test and benchmark the performance of our RUL prediction model are presented in Section III. The computational results are discussed in Section IV. Section V concludes this work.

II. Deep Learning-Nonstationary Gaussian Process Regression RUL Prediction Model

In this section, we first provide an overview of our deep learning-nonstationary Gaussian process regression model for RUL prediction and prognostic. We will refer to our model with the acronym DL-NSGPR. We then discuss the details of the main elements in this model.

*A. Overview of Our RUL Prediction Model*

We begin with a high-level overview of the RUL model and leave the details to the following subsections. The general architecture of the model is shown in Fig. 1.

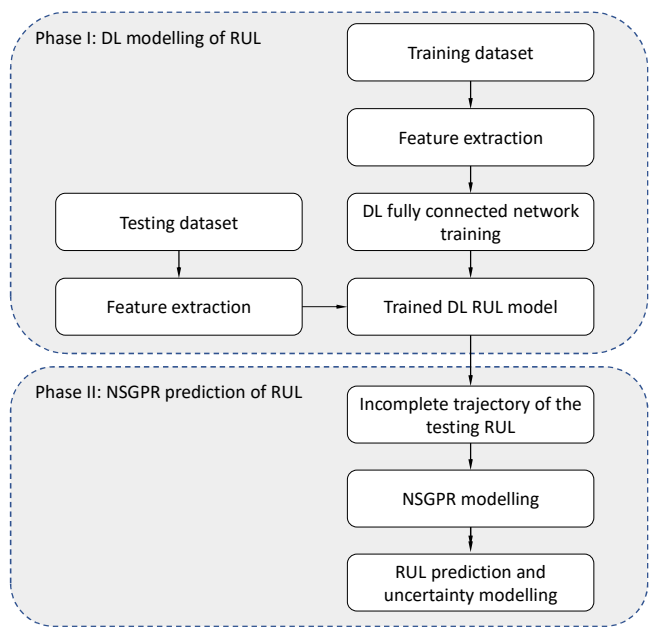

Fig. 1. Overview of the DL-NSGPR model for RUL prediction

This DL-NSGPR model for RUL prediction consists of two phases, a DL modeling phase, and a nonstationary GPR prediction phase. In Phase I, referred to as the DL modelling of RUL, we first conduct feature extraction from the input training data, which consists of a group of run-to-failure trajectories from the sensors data (i.e., up to the time to failure $t_f$). The run-to-failure enables the calculation of complete RUL trajectories up to $t_f$ in the training dataset (details in Section III). The purpose of this feature extraction step (details in Section II.B) is to combine different variables to reduce the dimensionality of the input training dataset. We then use the features extracted to train the fully connected DL network for the RUL estimation (details in Section II.D). Note that the trained DL model at the end of Phase I predicts the RUL during system operation at every point in time $t < t_f$ when sensor data is available. In the testing dataset, which contains incomplete degradation trajectories (i.e., truncated before machine failure), the DL updates its RUL prediction up to the time when the sensor data is truncated or censored $t_c$. For clarity, we index $RUL_t$ with the time at which the prediction is carried out. This concludes Phase I of the DL-NSGPR model.

There are two motivations for appending Phase II to this intermediate output. The first is a conceptual consideration and consists in the fact that the Phase I DL prediction is deterministic and lacks uncertainty estimation. The second is a practical consideration and consists in the fact that the DL $RUL_t$ predictions exhibit oscillations of varying amplitudes over time, and this ultimately degrades the accuracy of the prediction. To achieve a stochastic and more accurate RUL prediction, we append Phase II and its NSGPR to our model. We will discuss in detail in datasets used in this work in Section III and the results in Section IV, but for the time being, we provide Fig. 2 to illustrate these motivations and the differences in outputs between the Phase I and Phase II RUL predictions for engine No. 1 in the testing dataset. We see in Fig. 2 that the NSGPR significantly reduces the oscillations of the RUL prediction compared with the DL outputs.

The DL model from Phase I predicts the $RUL_t$ trajectory from $t_0$ up to $t_c$ in the testing dataset, and this trajectory is fed to the NSGPR in Phase II. We use the NSGPR to model the dynamics of these incomplete $RUL_t$ predictions in the testing dataset. By 'learning' from these predictions, the NSGPR captures the main trend in the RUL development over time. It is then used to provide more accurate and stochastic RUL predictions up to the time to failure along with uncertainty quantification around its prediction (in the form of confidence intervals for example. Details in Section II.D). The reader interested in a result illustrating this statement (before examining the technical details of the DL-NSGPR) may jump ahead to Section IV and glance at Fig. 7.

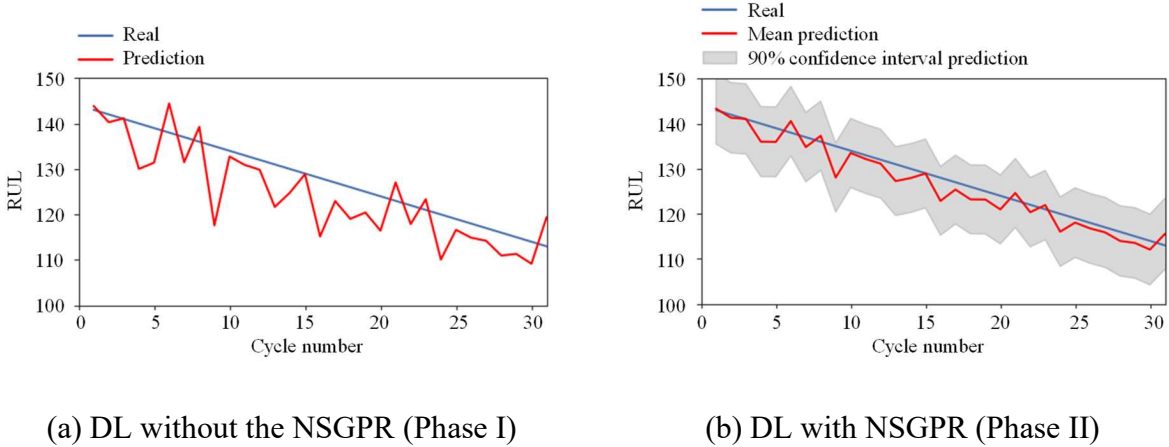

(a) DL without the NSGPR (Phase I)  (b) DL with NSGPR (Phase II)

Fig. 2. RUL prediction for engine No. 1 in the testing dataset

We designed this model to address the three challenges of data-driven RUL prediction discussed in the Introduction. In order to deal with the high dimensional input features, we leveraged the DL in Phase I because it is more powerful in dealing with high dimensional inputs compared with its shallow machine learning counterparts [30]. Furthermore, to analyze high dimensional features more effectively, we used a feature extraction method (details in Section III.B) to combine variables and remove others that do not contain signatures of the degradation process. To deal with feature noise, we used the NSGPR for the reasons discussed previously. Finally, in order to deal with the time dependence of the system degradation and RUL prediction, we added the time variable to the input vector of the DL network (details in Section III.C). We assume the training and testing machines follow similar degradations and failure modes. Embedding the time variable in the input vector of the DL model helps the network 'learn' the inherent time dependence of the degradation and RUL prediction. It is the totality of these choices and the integration of deep learning with non-stationary Gaussian process regression, that enables us to address the three challenges noted previously, to provide highly accurate RUL predictions with tight uncertainty bounds, as we will see in Section IV.

In the next subsections, we discuss the technical details of the different pieces in our DL-NSGPR RUL prediction model.

## B. Singular Value Decomposition Feature Extraction

As shown in Fig. 1, the first step in our model is to extract features from the training dataset in order to lower the dimensionality of the input data. We achieve this through singular value decomposition (SVD). In linear algebra, SVD is a particular factorization of a real or complex matrix [31]. The essence of SVD is that an $m \times n$ matrix $X$ can be decomposed into three matrices as shown in Eq. 1, where $\Sigma$ is a square diagonal matrix of size $r \times r$ with $r$ is the rank of $X$, $U$ is an $m \times r$ semi-unitary matrix, and $V$ is an $n \times r$ semi-unitary matrix (such that $U^T U = V^T V = I_{r \times r}$).

$$X = U\Sigma V^T \quad (1)$$

The left singular vectors $U$ can be calculated as a set of orthogonal eigenvectors of $XX^T$. The right singular vectors $V^T$ can be calculated by the eigenvectors of $X^T X$. The non-negative singular matrix $\Sigma$ is calculated by the square roots of the non-negative eigenvalues of $X^T X$ and $XX^T$ [31]. The entries of the $\Sigma$ are arranged in decreasing order ($\sigma_1 \geq \sigma_2 \geq \cdots \geq \sigma_r$).

Here, we leverage a truncated SVD method to compute a low dimensional representation of the original high dimensional input data and to extract the primary modes of the input features. The truncated SVD is shown in Eq. 2, where $U_t$ consists of the first $t < r$ column vectors of $U$, $V_t^T$ consists of the first $t < r$ row vectors of $V^T$ and $\Sigma_t$ consists of the $t$ largest singular values in $\Sigma$.

$$X_t = U_t \Sigma_t V_t^T \quad (2)$$

The rest of the original matrices is discarded. This method can provide a low dimensional representation of the full-dimensional SVD. The truncated SVD is no longer an exact decomposition of the original matrix, but the approximate $X_t$ is the closest approximation to $X$ that can be achieved by a matrix of rank $t$ as shown in Eq. 3.

$$X_t = argmin_{X^*} \sum_{i=1}^{m} \sum_{j=1}^{n} (X_{ij} - X_{ij}^*)^2 \qquad (3)$$

We use this truncated SVD feature extraction to reduce the high dimensional feature input of the dataset to a lower dimensional representation. We adopt a general rule discussed in [32] and given in Eq. 4 to determine the truncation number $t$.

$$\sum_{i=1}^{t} \sigma_i \geq 0.9 \sum_{j=1}^{r} \sigma_j \qquad (4)$$

We calculate the output of the feature extraction as a reduced dimension version of $X$ as $\tilde{X} = U_t \Sigma_t$ with a dimension of $m \times t$. This serves as part of the input of the DL network as discussed in detail in the next subsection.

*C. Deep Learning RUL Modelling*

A central element in our RUL prediction model is DL modelling. As shown in the top panel in Fig. 1, the input of the DL model is the result of the feature extraction step, and the output of the DL model is the RUL estimations used to build the NSGPR time streaming model. In this subsection, we discuss the details of this DL model.

The DL model consists of one input layer, three hidden layers ($H_1$, $H_2$, and $H_3$), and one output layer. The input layer consists of the results of SVD feature extraction and a time variable. Each hidden layer includes a Leaky ReLU activation function [33] after each linear neuron operation. The expression of the Leaky ReLU activation function is provided in Eq. 5. The output of this DL model is the RUL estimation. The schematic of the DL model is provided in Fig. 3.

$$f(x) = \begin{cases} -\alpha x & for\ x < 0 \\ x & for\ x \geq 0 \end{cases} \tag{5}$$

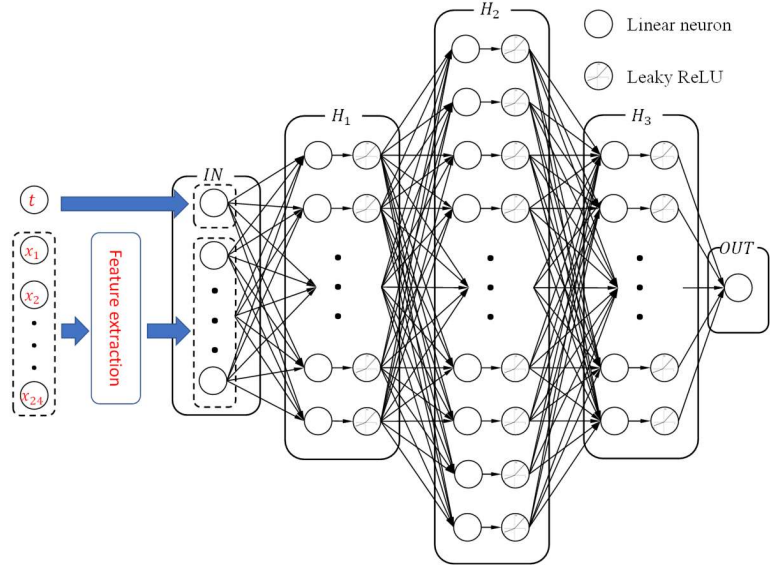

Fig. 3. Schematic of the fully connected DL neural network for RUL estimation

We implement the $l_2$ regularization in the backpropagation step to avoid overfitting and sharp decision boundaries in the feature space. We use the adaptive moment estimation (Adam) to obtain the optimal weight function [34]. The parameters of the DL network are the weight functions, which are trained automatically by Adam based on the training dataset. The hyperparameters of the DL model include the optimizer learning rate ($lr$), the $l_2$ regularization parameter ($\lambda$), the leaky ReLU parameter ($\alpha$), and the hidden layer size ($H_1, H_2, H_3$), $\boldsymbol{h} = [lr, \lambda, \alpha, H_1, H_2, H_3]$. We set the learning rate $lr = 5 \times 10^{-4}$, $l_2$ regularization $\lambda = 10^{-3}$, leaky ReLU parameter $\alpha = 0.2$, and $H_1 = H_3 = 50, H_2 = 100$ in our computational experiments obtained by trial and error. This selection of hyperparameters provides superior performance in our computational experiments. That being said, we recognize these hyperparameters can be further optimized, and we leave this as fruitful venue for future works.

D. *Non-stationary Gaussian Process Regression RUL Prediction*

The reasons for appending Phase II and the NSGPR to our RUL prediction model were discussed in Section II. A and illustrated in Fig. 2. The Gaussian process regression (GPR) is a Bayesian approach that can be used to model time streaming data. As shown in Phase II in Fig. 1, the GPR surrogate model is used here to model the RUL evolution in time, which in effect is used to predict the failure time of the machine. A brief introduction to GPR and our nonstationary kernel upgrade are provided next. For more details, the reader is referred to [35].

Consider $m$ realizations of a time streaming dataset in $m$ time steps. The input variables are denoted as $X_s = (x_1, x_2, \ldots, x_m)$, and each $x_i$ is a $d$ dimensional vector. The values of the output objective functions are denoted as $y_s = (y_1, y_2, \ldots, y_m)$, with each $y_i = Y(x_i)$, as the $m$ realizations of a stochastic function $Y(x)$, say RUL predictions. The GPR is formulated in Eq. 6, with $Y(x)$ decomposed into a deterministic mean approximation $f(x)$, and a stochastic part $Z(x)$. The stochastic part is a centered Gaussian process characterized by its covariance function. A quadratic function, as shown in Eq. 7, is used in this work to model the mean part $f(x)$, where $\beta$ is the weight and bias matrix. The stochastic part covariance function ($Cov(Z(x_i), Z(x_j))$) is modeled by selected kernel functions, and we will discuss our choice of covariance functions shortly. With the mean regression, covariance kernel functions, and $Y(x)$ decomposition formulation, the GPR prediction of new input ($x_{new}$) is considered as a posterior Gaussian distribution, which is shown in Eq 8. The Gaussian distribution mean and variance values are calculated by Eq. 8, where $\sigma_e$ is the Gaussian modeled noise standard deviation.

$$Y(x) = f(x) + Z(x) \tag{6}$$

$$\begin{cases} f(x) = H(x) \times \beta \\ \quad \text{with} \\ H(x) = [1, x, x^2] \\ x = [x_1, x_2, \ldots, x_d] \\ x^2 = [x_1^2, x_2^2, \ldots, x_d^2] \end{cases} \tag{7}$$

$$\mu_{new} = Cov(Z(X_s), Z(x_{new}))(Cov(Z(X_s), Z(X_s)) + \sigma_e^2 I)^{-1} \times$$
$$(Y(X_s) - f(X_s)) + f(x_{new})$$
$$\sigma_{new}^2 = Cov(Z(x_{new}), Z(x_{new}))Cov(Z(x_{new}), Z(X_s)) \times$$
$$(Cov(Z(X_s), Z(X_s)) + \sigma_e^2 I)^{-1} Cov(Z(X_s), Z(x_{new}))$$
(8)

Our GPR covariance function consists of two parts: (1) a stationary squared exponential kernel function, and (2) a non-stationary dot product kernel function. For the first part, we select a widely used stationary kernel, the squared exponential kernel [35], as shown in Eq. 9. For the second part, we select a simple non-stationary kernel, dot product, to upgrade our GPR time streaming modeling capability to a non-stationary version. According to [28], the stationarity of the covariance kernel functions stands for the behavior regarding the separation of any two points, say $x_i - x_j$ ($x_i, x_j \in X$). In contrast, the non-stationary kernel functions depend on the actual position of the points of $x_i$ and $x_j$ ($x_i, x_j \in X$). We apply a nonstationary kernel, dot product function (Eq. 10) [36], for the non-stationary part. We compare the prediction accuracy with a stationary versus a non-stationary GPR to demonstrate the latter's effectiveness in Appendix B. Our results show that the non-stationary kernel in the GPR significantly improves the prediction accuracy.

$$Cov_1(Z(x_i), Z(x_j)) = \sigma^2 R(x_i, x_j)$$
$$= \sigma^2 \exp\left(-\sum_{k=1}^{d} \theta_k (x_{ik} - x_{jk})^2\right)$$
(9)

$$Cov_2(Z(x_i), Z(x_j)) = \sigma_0^2 + x_i x_j \tag{10}$$

We add this non-stationary covariance $Cov_2$ to the stationary one $Cov_1$ for the overall covariance function in the NSGPR as Eq. 11.

$$Cov\left(Z(\pmb{x}_i), Z(\pmb{x}_j)\right) = Cov_1\left(Z(\pmb{x}_i), Z(\pmb{x}_j)\right) + Cov_2\left(Z(\pmb{x}_i), Z(\pmb{x}_j)\right) \tag{11}$$

We use this NSGPR to achieve a more accurate and stochastic RUL prediction [37]. In the next sections, we discuss the datasets used in this work, the computational experiments, and the results of our model. We also compare and benchmark our model predictions against those by other best-in-class models currently available.

### III. Datasets

To evaluate the performance of our DL-NSGPR model, we design computational experiments based on the turbofan engine dataset in the NASA prognostic repository [38, 39]. In this section, we introduce the details of this prognostic dataset and discuss the performance metric used to evaluate the RUL prediction accuracy.

This turbofan engine prognostic dataset [38, 39] consists of multivariate time series signals (in the unit of cycle numbers [38, 39]) collected from an engine dynamic simulation process. The engine run-to-failure simulations are carried out using C-MAPSS, a widely used simulation program for the transient operation of modern commercial turbofan engines. Two hundred engines' run-to-failure time series trajectories (dataset FD001) are used in our computational experiments. Each engine starts with different degrees of initial unknown wear and manufacturing variability. These engines operate normally at the start of each time series and develop a fault at some point during operation. The fault and degradation grow in magnitude until the system failure occurs. This data consists of two datasets: (1) a dataset with 100 engines for which whole trajectories from start to the engine failure is available; and (2) a testing dataset with another 100 engines with sensor trajectories truncated prior to the failure, as well as the ground truths for the corresponding RUL at the end of the incomplete trajectory.

The DL-NSGPR model is trained by the features and RUL trajectories of the engines in the training dataset, and it is used as described in the previous section to predict the RUL of the engines in the testing dataset. We then compare these model predictions with the ground truths of the RUL in the testing dataset to assess the accuracy of the predictions. Each training data record for a single engine in a run-to-failure trajectory is a 24-element vector, which consists of operational settings and noisy sensor measurements of the engine condition. These 24 input features are contaminated with noise. All failures are caused by the High-Pressure Compressor (HPC) in these 200 engines. For more details about the dataset, the reader is referred to [38, 39].

To illustrate what the data actually looks like, we provide next the 24-element feature vector of an engine from the training dataset, followed by the 24-element feature vector of an engine from the testing dataset.

We use engine No. 1 from the training set to illustrate the details of the whole run-to-failure trajectories of the 24 features and actual RUL with time and cycle number. The evolution of the 24 features are provided in Fig. 4, and the ground truth of the RUL development in Fig. 5.

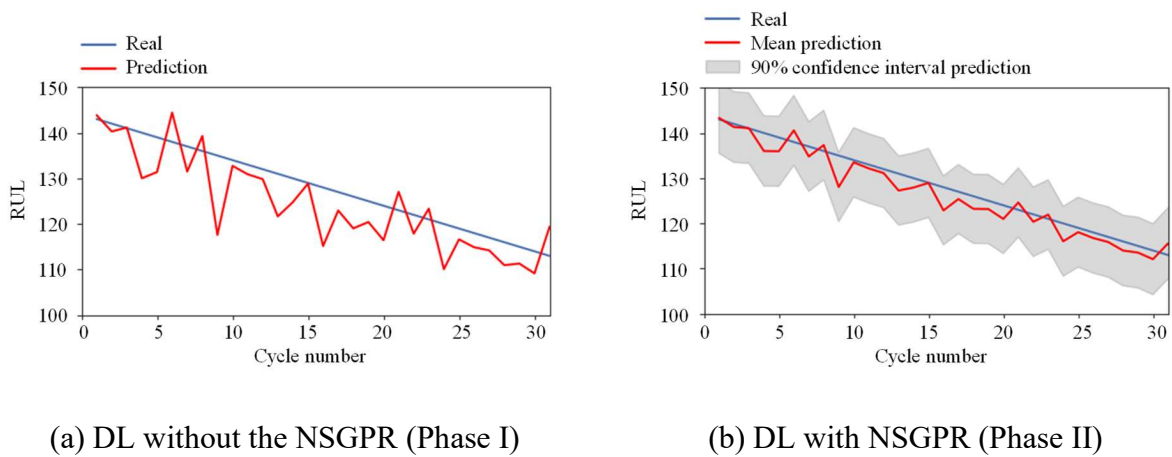

(a) DL without the NSGPR (Phase I)      (b) DL with NSGPR (Phase II)

Fig. 4. RUL prediction for engine No. 1 in the testing dataset

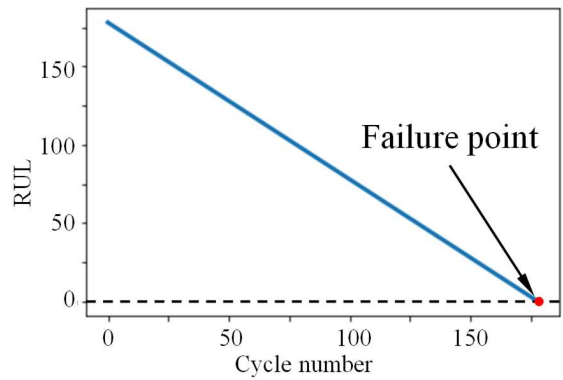

Fig. 5. RUL as a function of cycle number of engine No. 1 in the training dataset

Figure 4 shows that most features are quite noisy, which makes accurate RUL prediction challenging, and they exhibit different patterns of evolution over time (cycle number), some remaining roughly constant while others display exponential-like behavior, such as features 12 and 24 (increasing or decreasing) after a plateau. This can be indicative of an onset of a fault and the worsening of a degradation process [38, 39]. Other features exhibit no time variability, such as features 3 and 4, and carry no predictive value for this engine's RUL. The RUL ground truth in Fig. 5 is only available for the engines in the training datasets: knowing an engine has failed at, say 175 cycles, when it reached 100 cycles, its RUL was 75 and shrunk linearly with usage.

In Fig. 6, we show the available (incomplete) trajectories of the 24 features from another engine but this time from the testing dataset.

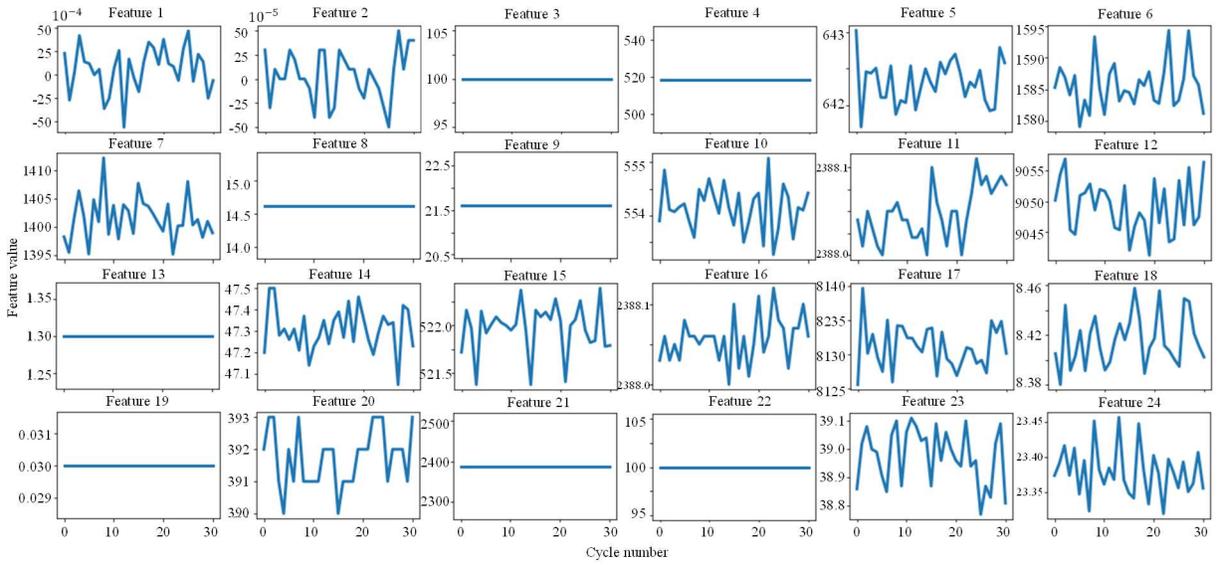

Fig. 6. The 24 features evolution with cycle number of engine No. 1 in the testing dataset

Notice how the trajectories in Fig. 6 are incomplete compared with those in Fig. 4. The former are truncated at an early stage during a sort of plateau period without a clear evolution trend. For example, feature 17 displays a clear exponential growth pattern in the training data (Fig. 4), whereas for the engine in the testing dataset (Fig. 6), the feature is noisy and exhibits no clear evolution pattern. This makes the RUL prediction for the engines in testing dataset considerably difficult [39]. These are precisely the challenges our DL-NSGPR is designed to tackle (high dimensional, noisy data, and early truncation).

Before getting to the results in the next section, we discuss the performance metric used to evaluate the RUL prediction accuracy of our DL-NSGPR prognostic model and others.

We use the DL-NSGPR model to predict the estimated value of the RUL of the 100 testing engines, $\tilde{y}_i$ ($1 \leq i \leq 100$). We then compare the estimated RUL at the time of feature truncation, $t_c$, with the ground truth $y_i$ ($1 \leq i \leq 100$) provided by the testing dataset. In order to compare the RUL prediction of our model with those of other models [18, 40, 41, 42, 43], we calculate a widely used residual index to measure the RUL prediction accuracy, the root mean square error (RMSE) as given in Eq. 12:

$$RMSE = (\sum_{i=1}^{100}(\tilde{y}_i - y_i)^2)/100 \qquad (12)$$

IV. Results and Discussion

In this section, we examine and discuss the computational results of our DL-NSGPR prediction of the RUL for the turbofan engines in the testing dataset. We first provide the results for a single engine example to illustrate the process and outcome of our model. We then provide the RUL predictions for the entire 100 engines in the testing dataset, and we compare the prediction accuracy of our model with that of other data-driven models [18, 40, 41, 42, 43].

*A. Results for Engine No. 1 in the Testing Dataset*

The incomplete feature trajectories of this engine were provided in Fig. 6. The feature data is only available up to cycle 31. Based on this data, and what the DL-NSGPR has 'learned' from the training dataset, our model predicts the mean RUL shown in Fig. 7 along with the uncertainty associated with it in the form of a 90% confidence interval.

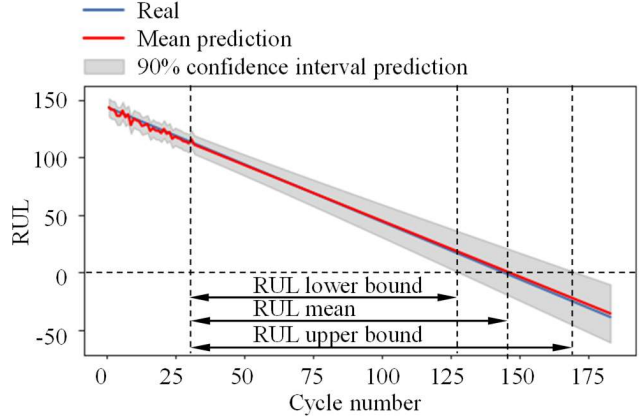

Fig. 7. The RUL estimation of the NSGPR model of the testing engine No. 1

At the time (cycle) when the feature data is no longer available (truncation time $t_c = 31\ cycles$), the DL-NSGPR estimates the engine's RUL to be $\tilde{y}_1 = 114\ cycles$. These are the estimated remaining cycles before the engine fails; failure is therefore expected at cycle 145. The ground truth for this engine's RUL at truncation time is $y_1 = 112\ cycles$. True failure therefore occurs at cycle 143. This difference is used to calculate the RMSE in Eq. 12 for all 100 engines in the testing dataset and to benchmark the performance of our predictive model against others. We note that this prediction is markedly accurate and the 90% confidence interval contains the ground truth, as shown in Fig. 7.

*B. RUL Predictions for the Entire Testing Dataset*

We provide in Fig. 8 the RUL predictions for the entire fleet of engines in the testing dataset. The figure also includes the confidence intervals and the ground truth for each engine. The results for engines No. 81 to 100 are provided in Table I and the remainders are included in Appendix C.

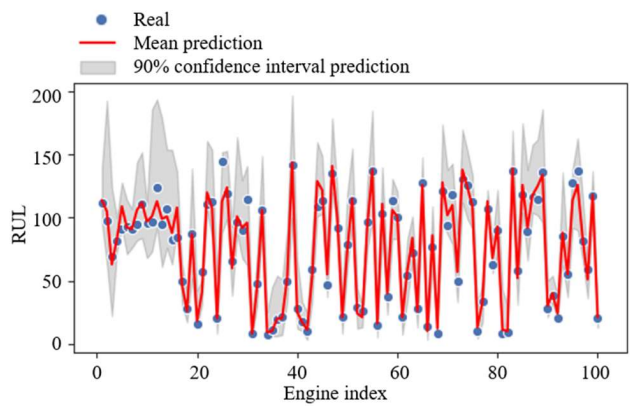

Fig. 8. RUL predictions for the entire 100 engines in the testing set

TABLE I. RUL PREDICTIONS, GROUND TRUTH VALUES, AND TRUNCATION TIME FOR TESTING ENGINES NO. 81-100

| Testing engine No. | RUL prediction at truncation time, $\tilde{y}_i$ | Ground truth RUL at truncation time, $y_i$ | Truncation time (cycle), $t_c$ |
|---|---|---|---|
| 81 | 11 | 8 | 213 |
| 82 | 10 | 9 | 162 |
| 83 | 139 | 137 | 73 |
| 84 | 52 | 58 | 172 |
| 85 | 126 | 118 | 34 |
| 86 | 92 | 89 | 110 |
| 87 | 118 | 116 | 56 |
| 88 | 125 | 115 | 68 |
| 89 | 134 | 136 | 177 |
| 90 | 30 | 28 | 146 |

| 91  | 40  | 38  | 234 |
| 92  | 24  | 20  | 150 |
| 93  | 88  | 85  | 244 |
| 94  | 55  | 55  | 133 |
| 95  | 114 | 128 | 89  |
| 96  | 126 | 137 | 97  |
| 97  | 84  | 82  | 134 |
| 98  | 51  | 59  | 121 |
| 99  | 118 | 117 | 97  |
| 100 | 21  | 20  | 198 |

Figure 8 indicates that the RUL predictions are accurate and within the vicinity of the ground truth values, all the latter are within the 90% confidence intervals. This suggests that the uncertainty quantification is valid and informative. The RMSE for all the testing engines is

$$RMSE_{DL-NSGPR} = (\sum_{i=1}^{100}(\tilde{y}_i - y_i)^2 = 7.4\ cycles$$

To demonstrate the RUL prediction performance at different levels of actual remaining life, the prediction development with actual remaining life is shown in Fig. 9. In addition, the prediction RMSE development at different actual RUL levels is shown in Fig. 10.

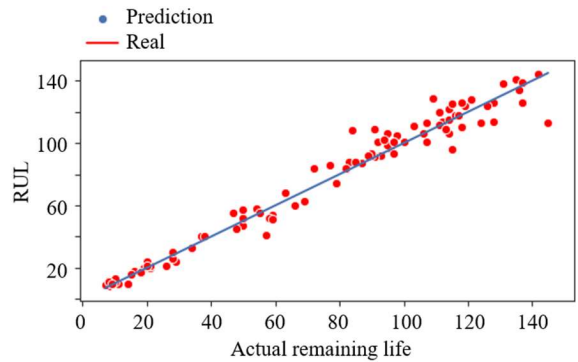

Fig. 9. RUL prediction development with actual remaining life

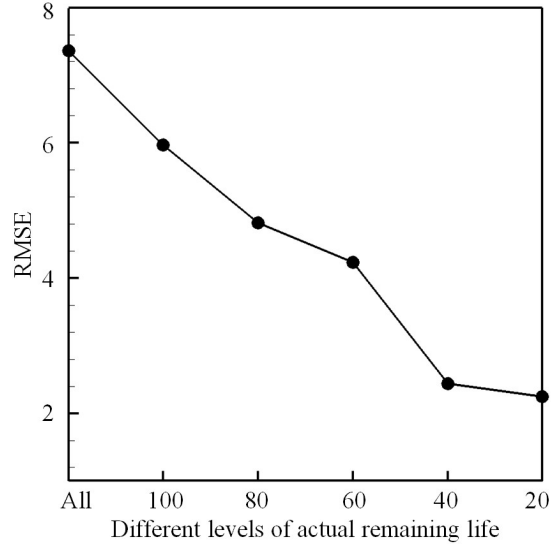

Fig. 10. RMSE development at different actual RUL levels

As shown in Fig. 9, the predictions are closer to the ground truth RUL curve when actual RUL is. In Fig. 10, the RMSE prediction accuracy improves as the actual RUL decreases. In other words, the closer the truncation time is to the actual machine failure, the more accurate the prediction is.

We show our RUL prediction and this RMSE are robust to the model initialization in Appendix D. How good or bad is this performance of the DL-NSGPR? To answer this question, it is useful to provide context and examine the performances of other best-in-class data-driven predictive models. These deterministic models include the support vector regression (SVR) [40], relevance vector regression (RVR) [40], multiple layer perceptron (MLP) [40], convolutional neural network (CNN) [40], deep Weibull recurrent neural network (DW-RNN) [41], multi-task learning recurrent neural network (MTL-RNN) [41], long short-term memory (LSTM) [42], and semi-supervised model [43]. We trained these models on the same training dataset and tested them on the same testing dataset as our DL-NSGPR. The results are provided in Table II.

TABLE II. COMPARISON OF RUL PREDICTION RMSE VALUE OF DL-NSGPR AND OTHER DATA-DRIVEN MODELS

| Data-driven model | RMSE | RMSE of *model i* over DL-NSGPR, $\frac{RMSE_{model\ i}}{RMSE_{DL-NSGPR}}$ |
|---|---|---|
| SVR [40] | 21.0 | 284% |
| RVR [40] | 23.8 | 322% |
| MLP [40] | 37.6 | 508% |
| CNN [40] | 18.5 | 250% |
| DW-RNN [41] | 22.5 | 304% |
| MTL-RNN [41] | 21.5 | 290% |
| LSTM [42] | 16.1 | 218% |
| Semi-supervised [43] | 12.56 | 170% |
| DL-NSGPR | 7.4 | — |

Table II indicates that the DL-NSGPR significantly outperforms other data-driven predictive models by a factor of 1.7 to 5.1. To the best of our knowledge, these are the most accurate deterministic RUL predictive models. The RUL prediction and the comparison between alternative methods in other C-MAPSS datasets (subsets FD002, FD003, and FD004) are provided in Appendix E. Our DL-NSGPR provides consistent and significant advantage over best-in-class alternatives for all these datasets in terms of RMSE and accuracy. In addition, we compare our DL-NSGPR with HI data fusion method [23] in Appendix F. The results show a significant advantage and dominating performance of our DL-NSGPR over HI method that DL-NSGPR has significantly smaller prediction error at all levels of actual RUL.

The reasons for this excellent performance of the DL-NSGPR are threefold. First, we used feature extraction to reduce the dimensionality of the 24-features input data and removed uninformative features, such as features 3 and 4 in Fig. 4. This produces more informative input data, and it helps build and train a more accurate model. Second, the DL neural network model is complex enough to capture the effects of the input features on the RUL estimation. The added time variable helps the DL model 'learn' the time dependency between system degradation and RUL. Finally, the NSGPR delivers on the advantages discussed in Section II. A (learning the dynamics of the incomplete *RUL* predictions in the testing dataset), and it ultimately provides more accurate and stochastic RUL predictions.

C. *Comparison Between DL-NSGPR with Stochastic RUL Prediction Model*

In the previous subsection, we benchmarked the accuracy of our (single point) mean RUL predictions against those of other deterministic models. In this subsection, we examine the uncertainty estimation performance of the DL-NSGPR. To this end, we compare the confidence intervals obtained with our model against another best-in-class stochastic RUL prediction model known as the non-homogeneous continuous-time hidden semi-Markov process (NHCTHSMP) [18]. For a fair comparison, we train and test the two models with the same datasets. More specifically, we train our model with the first 60 engines in the training set, as done in [18], and we test our model on engines No. 81-100 engines (20 engines), also as done in [18]. The feature trajectories for the testing engines in [18] are truncated at cycle number 50, and the same is done with the DL-NSGPR model in this subsection. We also use 95% confidence interval for the uncertainty quantification as done in [18]. The RUL predictions for the testing engines with both models along with the 95% confidence intervals are provided in Fig. 11.

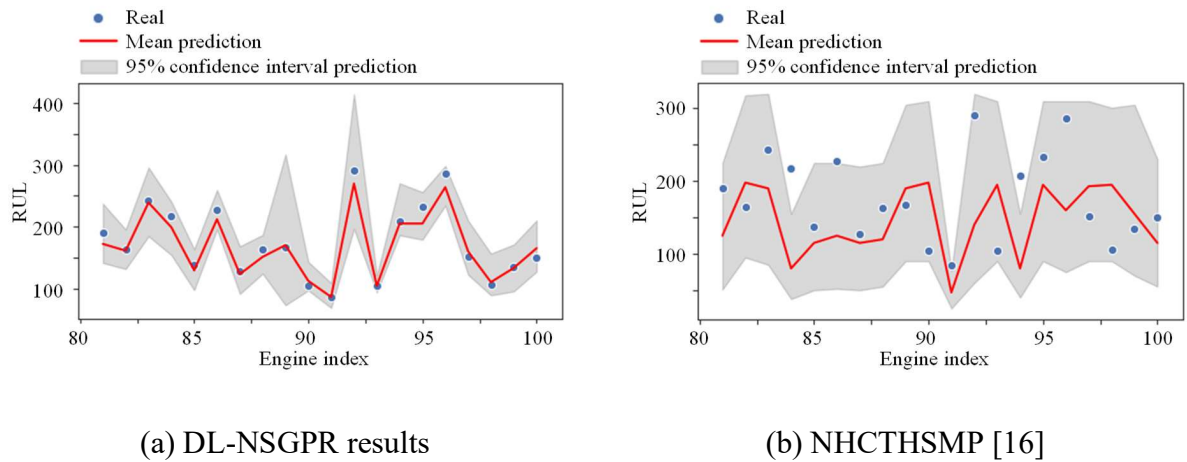

(a) DL-NSGPR results            (b) NHCTHSMP [16]

Fig. 11. The RUL prediction for engines No. 81-100 in the testing of the best-in-class condition-based prognostic models

Figure 11 shows that the DL-NSGPR outperforms the NHCTHSMP in this secondary test in terms of prediction accuracy. More specifically, we see that ground truth RUL in this testing dataset are significantly closer to the predictions with the DL-NSGPR than with the NHCTHSMP model. The testing RMSE of the mean RUL predictions with both models are:

$$RMSE_{NHCTHSMP} = 79.6 \ cycles$$

$$RMSE_{DL-NSGPR} = 12.8 \ cycles$$

The accuracy of the DL-NSGPR is about 6 times better than that of the NHCTHSMP model on this testing dataset (622%). This places the NHCTHSMP in roughly the same performance bin as that of the MPL relative to the DL-NSGPR (see Table II). Furthermore, beyond the accuracy of the mean RUL prediction, we note that the confidence intervals with both models are significantly different, the NHCTHSMP exhibiting a much larger spread than the DL-NSGPR. To compare the uncertainty estimation with both models, we calculate the average size of the confidence intervals on all testing engines. The results are provided in Fig. 12.

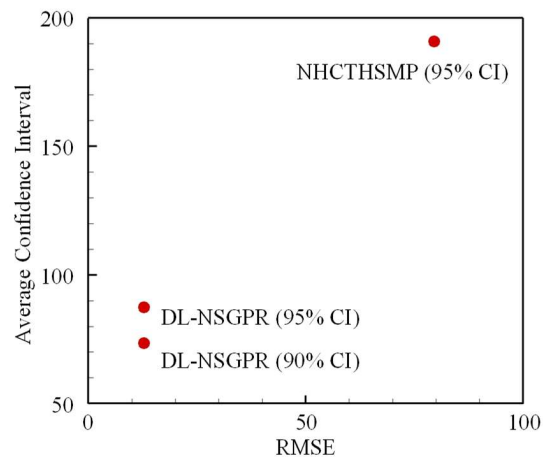

Fig. 12. Comparison of RMSE and average confidence interval between NHCTHSMP [18] and our DL-NSGPR

Figure 12 demonstrates a clear advantage of the DL-NSGPR over the NHCTHSMP, both in terms of improved mean RUL prediction accuracy and tighter uncertainty bounds.

We conclude this section with the hope and expectation that, just like our model outperformed other previous ones, future models will be proposed that will undoubtedly outperform ours. This dialectic and benchmark setting is healthy and useful for our community, and it will help advance the system reliability and PHM agenda. We suggest that the two axes in Fig. 12 or similar measures for prediction accuracy and uncertainty quantification be jointly used to assess future models.

## V. Conclusion

In this work, we addressed three challenges in data-driven RUL prediction, namely the handling of high dimensional input features, the robustness to noise in sensor data and prognostic datasets, and the capturing of non-stationarity or time-dependency of system degradation and RUL prediction given sensor data. We devised a highly accurate RUL prediction model with uncertainty quantification, which integrates and leverages the advantages of deep learning and nonstationary Gaussian process regression (DL-NSGPR). We examined and benchmarked our model against other advanced data-driven RUL prediction models using the turbofan engine dataset from the NASA prognostic repository. Our computational experiments showed that: (1) the DL-NSGPR predictions are highly accurate with RMSE 2 to 6 times smaller than those by competing models; (2) the RUL uncertainty bounds are both valid and significantly tighter than those by another best-in-class stochastic prediction model.

This work should be considered in light of its limitations, and these constitute fruitful venues for future work. First, our RUL prediction model was only validated using the turbofan engine prognostic dataset; more extensive testing will be conducted in the future to further challenge or confirm its advantages in other applications. Second, the hyperparameters of the DL network adopted in this work were selected based on heuristics and trial-and-error; they can be more systematically optimized in the future to achieve even better RUL prediction and uncertainty quantification. Third, other (unsteady) kernels for the NSGPR than the one used here can be examined in the future and their contributions to the overall model performance assessed. Fourth, other end-to-end machine learning methods and advanced Gaussian process models, such as deep

Gaussian process [44], can be examined in the future and compared with our DL-NSGPR model for RUL prediction.

APPENDIX

*A. Examining the Effectiveness of NSGPR*

In this appendix, we examine the effectiveness of the NSGPR in our RUL prediction model by comparing the RUL prediction with and without NSGPR. The RUL predictions without NSGPR are taken from the last step prediction of the DL network in the testing dataset. The RUL prediction without NSGPR the testing engine No. 1 is shown in Fig. A. 1.

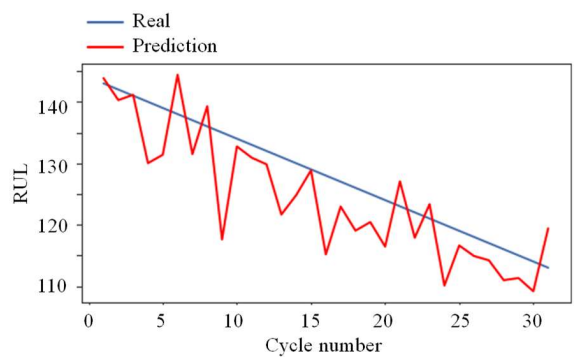

Fig. A.1. The RUL estimation of the DL network of the testing engine No. 1

First, we note that the RUL estimation curve is oscillating compared with the RUL prediction with NSGPR. Second, the RUL prediction is 120, which has a larger error than that of the prediction with NSGPR. The RUL predictions without NSGPR of 100 overall testing engines are shown in Fig. A.2.

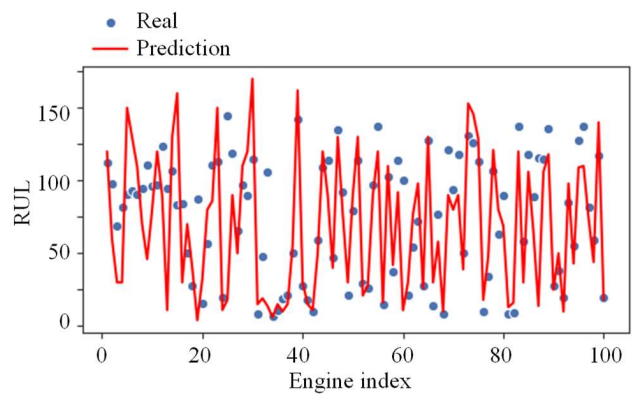

Fig. A.2. The RUL estimation of the entire 100 engines in the testing set of DL network

First, the RUL prediction in Fig. A. 2, does not provide the lower and upper bounds, and the use of NSGPR supports our model with uncertainty quantification. Second, the RMSE of the RUL prediction without NSGPR is 32.8. The use of RMSE improves the prediction accuracy significantly by decreasing RMSE to 7.4. The RUL prediction without NSGPR for the secondary computational experiments in Section IV.C is shown in Fig. A. 3.

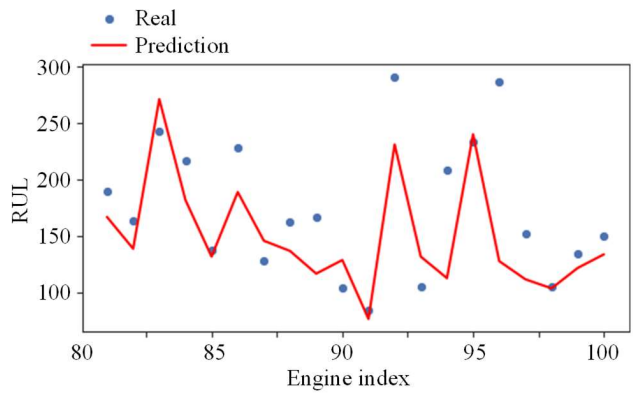

Fig. A.3. The RUL prediction at cycle number 50 for engine No. 81-100 of the DL RUL estimation model

Similar to the results of 100 testing engines RUL prediction, the use of NSGPR in this test improves the RUL prediction and provides uncertainty quantifications. The RMSE of the prediction improves from 49.6 to 12.8.

B. *Effectiveness Test of the Non-Stationary Kernel in GPR*

In this Appendix, we test the effectiveness of the non-stationary kernel in the GPR model and our RUL prediction. We compare the RMSE index of the RUL prediction in all subsets shown in Table E.1 for DL without GPR, with a stationary GPR, and with the NSGPR in Table B. I.

TABLE B. I. RMSE OF THE RUL PREDICTION OF DL WITHOUT GPR, GPR, AND NSGPR

| Dataset | FD001 | FD002 | FD003 | FD004 |
|---|---|---|---|---|
| DL without GPR | 32.8 | 69.5 | 33.2 | 58.7 |
| GPR | 10.9 | 19.6 | 11.3 | 12.5 |
| NSGPR | 7.4 | 11.8 | 7.5 | 8.3 |

Table B. I clearly illustrates the effectiveness and importance of adopting a non-stationary kernel in the GPR to improve the RUL prediction accuracy.

*C. RUL Predictions and Uncertainty Estimations for the Entire Testing Dataset*

TABLE C.I. RUL PREDICTIONS, GROUND TRUTH VALUES, AND TRUNCATION TIME FOR 100 TESTING ENGINES

| Testing engine No. | RUL prediction, $\tilde{y}_i$ | Ground truth RUL, $y_i$ | 90% confidence interval, [lb; ub] | Truncation time (cycle), $t_c$ |
|---|---|---|---|---|
| 1 | 114 | 112 | [96; 139] | 31 |
| 2 | 105 | 98 | [63; 193] | 49 |
| 3 | 63 | 69 | [21; 124] | 126 |
| 4 | 83 | 82 | [73; 100] | 106 |
| 5 | 109 | 91 | [92; 128] | 98 |
| 6 | 92 | 93 | [71; 114] | 105 |
| 7 | 91 | 91 | [76; 107] | 160 |
| 8 | 106 | 95 | [82; 144] | 166 |
| 9 | 112 | 111 | [84; 152] | 55 |
| 10 | 97 | 96 | [68; 111] | 192 |
| 11 | 101 | 97 | [73; 186] | 83 |
| 12 | 113 | 124 | [92; 194] | 217 |
| 13 | 99 | 95 | [61; 179] | 195 |
| 14 | 101 | 107 | [76; 154] | 46 |
| 15 | 88 | 83 | [63; 154] | 76 |
| 16 | 108 | 84 | [89; 137] | 113 |
| 17 | 47 | 50 | [31; 64] | 165 |
| 18 | 26 | 28 | [17; 41] | 133 |
| 19 | 87 | 87 | [52; 139] | 135 |
| 20 | 18 | 16 | [13; 42] | 184 |
| 21 | 41 | 57 | [31; 51] | 148 |
| 22 | 120 | 111 | [105; 161] | 39 |
| 23 | 108 | 113 | [84; 154] | 130 |
| 24 | 20 | 20 | [8; 51] | 186 |
| 25 | 113 | 145 | [88; 152] | 48 |
| 26 | 124 | 119 | [99; 153] | 76 |
| 27 | 60 | 66 | [38; 93] | 140 |
| 28 | 101 | 97 | [87; 164] | 158 |
| 29 | 92 | 90 | [68; 130] | 171 |
| 30 | 96 | 115 | [62; 141] | 143 |
| 31 | 8 | 8 | [5; 19] | 196 |
| 32 | 45 | 48 | [30; 72] | 145 |
| 33 | 106 | 106 | [87; 149] | 50 |
| 34 | 9 | 7 | [3; 25] | 203 |
| 35 | 10 | 11 | [6; 45] | 198 |
| 36 | 20 | 19 | [6; 54] | 126 |

| 37 | 20 | 21 | [7; 51] | 121 |
| --- | --- | --- | --- | --- |
| 38 | 52 | 50 | [33; 91] | 125 |
| 39 | 144 | 142 | [100; 197] | 37 |
| 40 | 25 | 28 | [12; 67] | 133 |
| 41 | 17 | 18 | [10; 33] | 123 |
| 42 | 11 | 10 | [3; 25] | 156 |
| 43 | 54 | 59 | [34; 86] | 172 |
| 44 | 129 | 109 | [101; 160] | 54 |
| 45 | 122 | 114 | [103; 149] | 152 |
| 46 | 55 | 47 | [37; 76] | 146 |
| 47 | 141 | 135 | [120; 179] | 73 |
| 48 | 101 | 92 | [89; 123] | 78 |
| 49 | 21 | 21 | [8; 41] | 303 |
| 50 | 74 | 79 | [58; 119] | 74 |
| 51 | 115 | 114 | [95; 155] | 144 |
| 52 | 24 | 29 | [12; 44] | 189 |
| 53 | 21 | 26 | [13; 31] | 164 |
| 54 | 93 | 97 | [74; 127] | 121 |
| 55 | 138 | 137 | [99; 185] | 113 |
| 56 | 16 | 15 | [5; 31] | 136 |
| 57 | 111 | 103 | [87; 140] | 160 |
| 58 | 40 | 37 | [23; 59] | 176 |
| 59 | 106 | 114 | [83; 131] | 94 |
| 60 | 101 | 100 | [83; 121] | 147 |
| 61 | 21 | 21 | [7; 35] | 159 |
| 62 | 58 | 54 | [34; 83] | 232 |
| 63 | 84 | 72 | [66; 101] | 155 |
| 64 | 26 | 28 | [13; 49] | 168 |
| 65 | 126 | 128 | [97; 148] | 71 |
| 66 | 10 | 14 | [3; 19] | 147 |
| 67 | 86 | 77 | [62; 109] | 71 |
| 68 | 12 | 8 | [10; 16] | 187 |
| 69 | 128 | 121 | [96; 174] | 54 |
| 70 | 102 | 94 | [87; 138] | 152 |
| 71 | 110 | 118 | [79; 143] | 68 |
| 72 | 57 | 50 | [33; 87] | 131 |
| 73 | 138 | 131 | [112; 170] | 112 |
| 74 | 124 | 126 | [99; 153] | 137 |
| 75 | 109 | 113 | [91; 134] | 88 |
| 76 | 13 | 10 | [4; 31] | 205 |
| 77 | 33 | 34 | [17; 49] | 162 |
| 78 | 113 | 107 | [106; 122] | 72 |
| 79 | 68 | 63 | [56; 83] | 101 |
| 80 | 93 | 90 | [71; 122] | 133 |
| 81 | 11 | 8 | [6; 43] | 213 |
| 82 | 10 | 9 | [5; 30] | 162 |
| 83 | 139 | 137 | [119; 169] | 73 |
| 84 | 52 | 58 | [31; 73] | 172 |

| 85 | 126 | 118 | [104; 175] | 34 |
| 86 | 92 | 89 | [66; 138] | 110 |
| 87 | 118 | 116 | [104; 158] | 56 |
| 88 | 125 | 115 | [102; 162] | 68 |
| 89 | 134 | 136 | [96; 186] | 177 |
| 90 | 30 | 28 | [21; 50] | 146 |
| 91 | 40 | 38 | [32; 54] | 234 |
| 92 | 24 | 20 | [18; 39] | 150 |
| 93 | 88 | 85 | [62; 136] | 244 |
| 94 | 55 | 55 | [37; 72] | 133 |
| 95 | 114 | 128 | [92; 139] | 89 |
| 96 | 126 | 137 | [90; 163] | 97 |
| 97 | 84 | 82 | [66; 117] | 134 |
| 98 | 51 | 59 | [36; 66] | 121 |
| 99 | 118 | 117 | [104; 137] | 97 |
| 100 | 21 | 20 | [13; 30] | 198 |

*D. Robustness Test of Model Initialization*

In this Appendix, we test the robustness of the RUL prediction to the model initialization. We independently conduct 10 trials of RUL prediction in our computational experiment by varying the random seed, and we provide the RMSE results in Fig. D. 1.

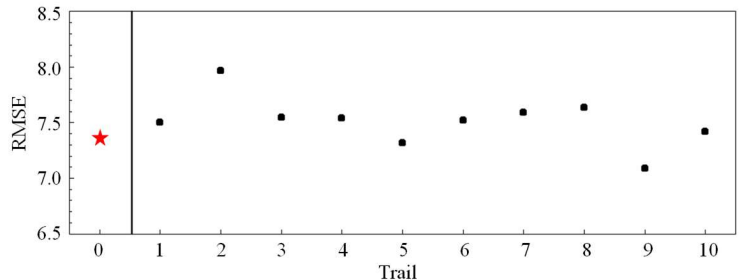

Fig. D.1. The RUL prediction robustness test to the model initialization

The RUL prediction in Section IV is labeled as the star sign of trail 0 with RMSE 7.4. The mean RMSE of the trail 1 to 10 is 7.5, and the standard deviation is 0.23. The standard deviation is 3% of the mean of the RMSE in the 10 independent trials. This small standard deviation shows that the RUL predictions are robust to the DL model initialization. In addition, the result shown in Section IV and Fig. 8 remains within 1.5% of the mean value in these 10 trails.

*E. RUL Prediction of All Engine Prognostic Subsets*

In this Appendix, we provide the RUL prediction accuracy over all the engine prognostic dataset (Table E.I). The results for FD002, FD003, and FD004 are provided in Table E. II, E. III, and E. IV, respectively.

TABLE E. I. DATASET INTRODUCTION

| Dataset | FD001 | FD002 | FD003 | FD004 |
|---|---|---|---|---|
| Training trajectories | 100 | 260 | 100 | 249 |
| Testing trajectories | 100 | 259 | 100 | 248 |
| Operating conditions | 1 | 6 | 1 | 6 |
| Fault conditions | 1 | 1 | 2 | 2 |

TABLE E. II. COMPARISON OF RUL PREDICTION RMSE VALUE OF DL-NSGPR AND OTHER DATA-DRIVEN MODELS IN FD002 SUBSET

| Data-driven model | RMSE | RMSE of *model i* over DL-NSGPR, $\frac{RMSE_{model\ i}}{RMSE_{DL-NSGPR}}$ |
| --- | --- | --- |
| SVR [40] | 42.0 | 356% |
| RVR [40] | 31.3 | 265% |
| MLP [40] | 80.0 | 678% |
| CNN [40] | 30.3 | 257% |
| DW-RNN [41] | 25.9 | 219% |
| MTL-RNN [41] | 25.8 | 219% |
| LSTM [42] | 24.5 | 208% |
| Semi-supervised [43] | 22.7 | 192% |
| DL-NSGPR | 11.8 | — |

TABLE E. III. COMPARISON OF RUL PREDICTION RMSE VALUE OF DL-NSGPR AND OTHER DATA-DRIVEN MODELS IN FD003 SUBSET

| Data-driven model | RMSE | RMSE of *model i* over DL-NSGPR, $\frac{RMSE_{model\ i}}{RMSE_{DL-NSGPR}}$ |
| --- | --- | --- |
| SVR [40] | 21.0 | 280% |
| RVR [40] | 22.4 | 298% |
| MLP [40] | 37.4 | 499% |
| CNN [40] | 19.8 | 264% |
| DW-RNN [41] | 18.8 | 251% |
| MTL-RNN [41] | 18.0 | 240% |
| LSTM [42] | 16.2 | 216% |
| Semi-supervised [43] | 12.1 | 161% |

| Data-driven model | | |
|---|---|---|
| DL-NSGPR | 7.5 | — |

TABLE E. IV. COMPARISON OF RUL PREDICTION RMSE VALUE OF DL-NSGPR AND OTHER DATA-DRIVEN MODELS IN FD004 SUBSET

| Data-driven model | RMSE | RMSE of $model\ i$ over DL-NSGPR, $\frac{RMSE_{model\ i}}{RMSE_{DL-NSGPR}}$ |
|---|---|---|
| SVR [40] | 45.3 | 546% |
| RVR [40] | 34.3 | 413% |
| MLP [40] | 77.4 | 933% |
| CNN [40] | 29.2 | 352% |
| DW-RNN [41] | 24.4 | 294% |
| MTL-RNN [41] | 22.8 | 275% |
| LSTM [42] | 28.2 | 340% |
| Semi-supervised [43] | 22.7 | 273% |
| DL-NSGPR | 8.3 | — |

From the results in Table E. II to E. IV, our DL-NSGPR can provide a consistent and significant advantage over best-in-class competitors in terms of RMSE and accuracy.

*F. Comparison with HI data fusion model*

In this Appendix, we compare our DL-NSGPR prediction accuracy with HI models Because Ref. [23] compares all HI related methods, including HI-non, HI-semi, HI-SNR, HI-kernel, and HI-DNN, and it demonstrates that the HI-DNN has the best performance among these methods, it is safe to state the superiority of DL-NSGPR over HI related methods in terms of estimation accuracy if it outperforms the HI-DNN model. We calculate the prediction error [23] with different levels of actual RUL and compare the results with those of HI-DNN as shown in Fig. F.1. The results show a significant advantage and dominating performance of the DL-NSGPR over the HI-DNN model: the former has significantly smaller prediction error compared with the latter at all levels of actual remaining life.

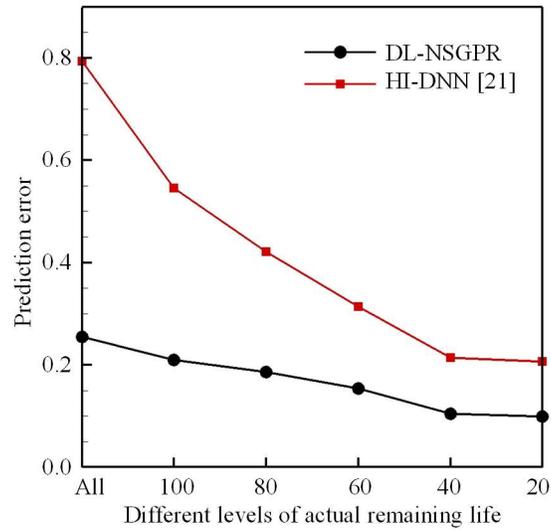

Fig. F.1. Prediction error comparison of the DL-NSGPR and HI-DNN [23] at different actual RUL levels

ACKNOWLEDGMENT

This work was supported in part by a Space Technology Research Institute grant from NASA's Space Technology Research Grants Program.